# Multi-Column RBF Neural Network Using Adaptive and Non-Adaptive Particle Swarm Optimization

Ammar O. Hoori, and Yuichi Motai, *Senior Member, IEEE*

**Abstract** The radial basis function neural network (RBFN) trained with a gradient descending algorithm provides an effective fully connected structure in both shallow and deep networks. The error correction (ErrCor), a state-of-the-art gradient-based training method, selects optimal hidden units to improve accuracy. Alternatively, as a population-based algorithm, the particle swarm optimization algorithm (PSO) uses the swarm experience to optimize RBFN parameters, offering global search and robustness to local minima. Adaptive PSO (APSO) has emerged as an improved variant of PSO. APSO algorithm improves convergence speed by dynamically adjusting swarm parameters during optimization. Both ErrCor and PSO demonstrate improved results and competitive convergence. However, with large datasets, these methods face scalability challenges such as excessive kernel computations and large hidden layer structures. A recent multi-column RBFN approach (MCRN) improves ErrCor performance by deploying small RBFNs in a parallel system. Inspired by MCRN's success, we propose two novel approaches to improve PSO performance: the multi-column RBFN with PSO (MC-PSO) and the multi-column RBFN with APSO (MC-APSO). These methods introduce parallel RBFN structures trained using evolutionary swarm methods. Each RBFN is independently trained on a specific spatial subset of the dataset using either PSO or APSO algorithms. These resulting specialist-trained RBFNs are tailored to their respective subsets. During testing, only selected RBFNs, where the test instance neighbors are located, contribute to the multi-column output. This specialization improves accuracy, while parallelism enhances speed. We evaluate the proposed methods on various benchmark datasets. The MC-PSO and MC-APSO outperform ErrCor, PSO, APSO, and MCRN in terms of accuracy and recall. They also demonstrate faster training and testing times in most experiments.

## I. INTRODUCTION[1]

THE radial basis function neural network (RBFN) showed a universal approximation ability that made it applicable to several shallow and deep learning research areas [1]–[5]. RBFN is a three-layered neural network (NN), where the Gaussian function (hence radial) is the hidden unit activation function. The number of hidden units, hidden weights, and their optimal parameters are the key factors in having a flawless RBFN [6]. A common approach for RBFN training involves using steepest descent to optimize parameters. For instance, both improved second-order (ISO) gradient method [6]–[8] and the error correction method (ErrCor) [9] incorporate second-order gradient information in their parameter updates. ISO uses a fixed number of hidden units, while ErrCor uses a step-by-step incremental hidden unit insertion. The number of hidden units incrementally increases with each epoch until convergence. Both methods show improved results compared to traditional gradient training methods. However, despite these improvements, prior methods still exhibit critical limitations. Gradient-based approaches such as ISO and ErrCor remain sensitive to local minima, require extensive gradient computations, and suffer from slow training when the hidden layer grows large [6], [9].

Meanwhile, particle swarm optimization (PSO) is a population-based technique that uses particle-like approach to converge towards a global minimum. The swarm particles cooperate in searching for an optimum solution by keeping track of each particle's previous best position and the best swarm particle position [10], [11]. Due to its success, PSO has been widely used in tuning RBFN during the past two decades [12]–[15]. PSO optimizes RBFN parameters by leveraging swarm intelligence, offering advantages such as global search capabilities and robustness to local minima. PSO stacks all the RBFN hidden units' parameters as the tunable elements of every single particle of the PSO swarm. Therefore, each particle has the parameters of the full-structured

Ammar O. Hoori is with the Department of Biomedical Engineering, Case Western Reserve University, Cleveland, OH 44106, USA (e-mail: ammar.hoori@case.edu).
Yuichi Motai is with the Department of Electrical and Computer Engineering, Virginia Commonwealth University, Richmond, VA 23284, USA (e-mail: ymotai@vcu.edu).

RBFN. With each training iteration, PSO updates the particles' parameters using previous swarm experience. In the fitness function minimization surface, the particles theoretically converge (fly) towards the best particle, which is the one with the optimum fitness value [16]. An adaptive PSO (APSO) was proposed to improve the PSO training performance [17]. It dynamically adjusts swarm parameters during the optimization process, enhancing convergence speed and accuracy while avoiding premature convergence, a common issue in standard PSO. Specifically, APSO increments/prunes the number of particles and tunes their speed inertia with each training iteration. However, APSO method requires a predefined up-limit with reserved memory storage space for the incremented particles. While each particle is a complete RBFN structure, the empty particle spaces are ineffectively reserved with each iteration until convergence [17]. Evolutionary methods like PSO and APSO, while offering global search, face slow convergence in high-dimensional parameter spaces and, in the case of APSO, inefficient memory usage due to reserved but unused particles [10], [11], [17].

Both gradient (ISO and ErrCor) and evolutionary (PSO and APSO) methods showed outstanding performance in terms of accuracy and convergence speed with small to medium size datasets. However, with extensive data, they suffer from excessive kernel computations, lack of accuracy, and slow convergence [17]. In solving these issues, a recent parallel ErrCor, dubbed multi-column RBFN (MCRN) [6], [8], was suggested to improve ErrCor accuracy and speed. MCRN employed smaller structure, specialized RBFNs in parallel, and ensemble results. The new multi-column method showed a promising way of chopping a dataset in feature-space into smaller subsets, training them using ErrCor individually, and ensemble the weighted-sum output. MCRN provides an improved solution to the gradient ErrCor method problem with a larger dataset [8].

In this paper, motivated by [6], we propose two multi-column RBFN with PSO (MC-PSO) and APSO (MC-APSO) methods, respectively. The new techniques use parallel stacked individual RBFNs. Each RBFN is trained from a chopped feature-space subset using evolutionary training PSO and APSO. The dataset is pre-chopped using a k-d tree algorithm. After training, each RBFN becomes a specialized RBFN for the trained subset region. With each new test input entry, the input vector is forwarded to a selected specialized RBFN based on the input neighbors' spatial locations. Only selected individual RBFNs will contribute to the network output. This will be further explained in subsection III. D.

The contributions of this paper are:

- Two novel MC-PSO and MC-APSO methods are introduced to improve PSO and APSO classification performance, respectively.
- Proposed methods are faster in training, especially with large datasets, compared to traditional evolutionary methods that suffer excessive computations.
- Dataset dividing into spatial subsets, and RBFNs training individually on each subset, eases the idea of parallelizing (multi-processing), a method appreciated in a deep learning environment.
- Methods are ready-parallel alternatives to the deep network final neural network fully connected layer.

The rest of this paper is organized as follows. The relevant studies of RBFN, PSO, and multi-column NNs are discussed in section II. Section III shows the proposed MC-PSO and MC-APSO in detail. Section IV presents and discusses the proposed methods' experimental results. The conclusion section is presented in Section V.

## II. Relevant Studies

This section demonstrates the literature reviews of the RBFN training, the use of the evolutionary PSO population method for RBFN training, and the parallel-wise idea of using a multi-column NNs structure.

### *A. RBFN Incremental Design, and Parameters Tuning*

The two primary undetermined settings in training are the number of hidden units and their parameters (center vector, width, and weight of each unit).

Several studies improved the gradient descent algorithms with different strategies to tune the RBFN parameters and optimize number of hidden units. Panchapakesan *et al.* [18] reduced the RBFN accuracy error by studying the effects of moving the centers and fine-tuning parameters on bounds for the gradient. Bruzzone and Prieto [19] used class memberships of training samples to select centers and widths of the hidden Gaussian functions of the RBFN classifier. Orr [20] proposed a regularized forward selection method that combined the forward subset selection and the zero-order regularization to select proper centers of RBFN units.

In terms of tuning the number of hidden units, a compact RBFN network design was suggested by Huang *et al.* [21] by introducing a generalized growing and pruning strategy, which Bortman and Aladjem [22] furtherly developed. Similarly, Wu *et al.* [23] proposed a self-adjustable number of RBFN hidden units for machine fault detection using self-organizing mapping. Xie *et al.* [7] proposed an improved second-order method (ISO) to tune the RBFN parameters. They used an improved Levenberg–Marquardt gradient training algorithm to overcome the significant second-order Jacobian matrix problem. Based on the work of Xie *et al.* [7], Yu *et al.* [9] introduced an incremental RBFN design by building a step-by-step RBFN structure using the error correction (ErrCor) algorithm. ErrCor calculates the root mean square error (RMSE) in each training step (epoch) and selects the input vector with the most error value as the newly inserted hidden unit to the RBFN structure.

### B. Evolutionary RBFN Training using PSO Algorithm

One of the successful techniques used to tune RBFN parameters and improve performance is swarm experience in convergence as used in the PSO algorithm. PSO provides a fast population-based method that can converge to best-tuned RBFN parameters.

Chen [24] proposed a PSO-aided orthogonal forward regression method to tune the RBFN parameters. Feng [25] proposed a self-generation RBFN trained with the PSO algorithm that adjusts the RBFN parameters while pruning the number of hidden units of the RBFN. Oh *et al.* [26] used the PSO and differential evolution to optimize the parameters of k-means clustering-based polynomial RBFNs. Han *et al.* [16] used an adaptive multi-objective gradient method with a PSO algorithm to tune the RBFN parameters and optimize the number of hidden units. Han *et al.* [17] designed a self-organizing radial basis function neural network (SORBF) using APSO. The APSO optimizes both RBFN parameters and the network size. Chen *et al.* [27] proposed an online modeling algorithm using RBFN. Each hidden unit had an adjustable center and covariance matrix, which were tuned using the PSO method. Chen *et al.* [28] also proposed a fast, novel adaptive RBFN for online system identification.

### C. Parallel Training Using Multi-Column RBFN

The idea of separating the dataset into small subsets and training individual NNs using these subsets has been widely promoted in ensemble learning literature. Recently, it has become even more applicable with the aid of using multi-node servers or multi-processing computers [6].

Shao *et al.* [29] proposed a multi-column NN that uses the idea of training multispectral NN as an alternative and efficient solution to using a single deep NN. Ciresan *et al.* [30] gathered and averaged the outputs of parallel NNs in a multi-column deep NN structure. Mall *et al.* [31] divided a large dataset into sparse subsets using KNN. Hoori and Motai [6] proposed a multi-column RBFN (MCRN). The MCRN differs from ensemble learning methods in that it divides each dataset into smaller subsets using the k-d tree algorithm and the way it selects the appropriate output from the small classifiers. The MCRN regional-specialized RBFNs improved accuracy and speed compared to a traditional-structured RBFN structure.

Inspired by the successful MCRN parallel training, we propose two novel multi-column RBFN, MC-PSO, and MC-APSO. The novel methods suggest the use of the PSO and the adaptive APSO as a training method to improve the internal parameters of each RBFN as will be explained in the following section.

## III. Multi-Column RBFN using PSO (MC-PSO) and APSO (MC-APSO) as Training Algorithm

This section constructs the two proposed multi-column methods in a step-by-step fashion. A quick spotlight on the idea of RBFN and recent gradient decent training algorithms is explained in subsection A. A recap of the PSO algorithm in tuning the RBFN parameters is detailed in subsection B, while subsection C illustrates the APSO algorithm in training RBFNs. Finally, the proposed MC-PSO and MC-APSO methods are explained in detail in subsection D.

### A. RBFN Fundamentals and Gradient Descent Training Algorithms

RBFN is used deliberately in many difficult applications for its efficient yet straightforward structure. The RBFN structure consists of three layers: The input layer, $\mathbf{x} = [x_1, x_2 \dots, x_i, \dots\ x_I]^T$; The hidden layer, $\mathbf{\Phi} = [\phi_1, \phi_2, \dots, \phi_h, \dots \phi_H]^T$; and a single output layer, $y$, as illustrated in Fig. 1. The hidden layer has Gaussian functions (1) as the activation functions of each hidden unit. The hidden activation function performs the inner product between any new input vector and the center vector by calculating the Euclidian distance between them and projecting the result in the Gaussian shape of that function [32]. The activation function is shown as:

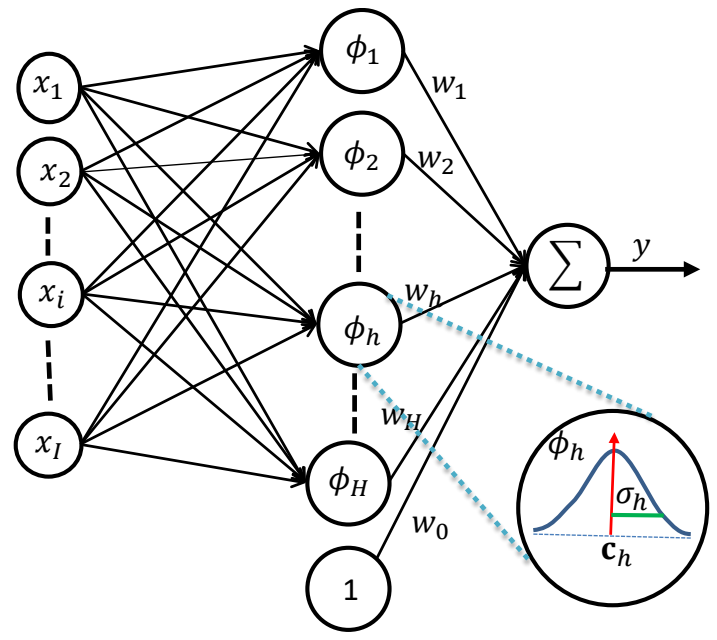

Fig. 1. The internal structure of a three-layered single-output RBFN with input vector $\mathbf{x}$ ; hidden layer $\mathbf{\Phi}$; and a single output, $y$. Each $h$ hidden layer unit has Gaussian function with $\mathbf{c}_h$ as a center vector and $\sigma_h$ as a width value.

$$\phi_h(\mathbf{x}) = \exp\left(-\frac{\|\mathbf{x} - \mathbf{c}_h\|^2}{\sigma_h^2}\right), \tag{1}$$

where the vector $\mathbf{c}_h$ and the value $\sigma_h$ are the center vector and width value of the $h$-th hidden unit $\phi_h$, respectively. The operation $\|.\|$ is the Euclidean distance. The NN output is a linear summation of the weighted hidden units, as shown:

$$y = \sum_{h=1}^{H} w_h \phi_h(\mathbf{x}) + w_0, \quad (2)$$

where $w_h$ represents the output weight value between the $h$-th hidden unit and the output, $y$, while $w_0$ represents the bias weight with an input 1 to the output unit for simplicity.

The RBFN convergence cost function is the root mean square error (RMSE) between the actual RBFN outputs and the desired outputs of the training vectors, as shown in (3):

$$E = \sqrt{\frac{1}{P}\sum_{p=1}^{P}(e^p)^2}\,, \qquad e^p = y_d^p - y^p, \quad (3)$$

where, $e^p$ is the error between $y_d^p$, the $p$-th desired output and its counterpart $y^p$, actual network output. $P$ is number of training dataset pairs $\{\mathbf{x}^p, y_d^p\}$, used to train RBFN.

A good training strategy is a crucial factor in obtaining good RBFN results [6], [9], [33]. The RBFN tunable parameters are the center vector, the width, and the output weights (including the bias) for each hidden unit. Moreover, a sufficient number of hidden units will increase the curvature representation of the output optimization cost function and hence the accuracy results of the RBFN.

The ISO [7] and the ErrCor [9], as two well-performed algorithms and successful gradient descent methods, are considered in this paper for comparison with other RBFN training methods. ISO, with a fixed number of hidden units, uses the gradient descent of each RBFN tunable parameter to update parameters until convergence, such that:

$$\frac{\partial e^p}{\partial w_h} = -\,\phi_h\left(\mathbf{x}_p\right), \qquad \frac{\partial e^p}{\partial w_0} = -1. \quad (4)$$

$$\frac{\partial e^p}{\partial c_{h,i}} = -\frac{2 w_h\, \phi_h\left(\mathbf{x}_p\right)\left(x_{p,i} - c_{h,i}\right)}{\sigma_h}. \quad (5)$$

$$\frac{\partial e^p}{\partial \sigma_h} = -\frac{w_h\, \phi_h\left(\mathbf{x}_p\right)\left\|\mathbf{x}_p - \mathbf{c}_h\right\|^2}{\sigma_h^2}. \quad (6)$$

The ErrCor method also uses the same gradient technique in tuning the parameters of the RBFN. However, unlike ISO, ErrCor starts with only one hidden unit and calculates the cost of RBFN for convergence. If convergence is not met, a new hidden unit is added iteratively until convergence. The new hidden center vector is carefully chosen among the training vectors. The selected vector showed the maximum violation error value when evaluating the training dataset for the current RBFN structure. This process of adding new hidden units continues until convergence. The incremental addition of ErrCor hidden units implies more accurate results and provides a smaller number of hidden units compared to ISO. Both methods use gradient descent to adjust network parameters. Nevertheless, the gradient descent methods are considered slow but successful iterative methods.

In our work, we also use two PSO population-based tuning RBFN methods: the PSO and the APSO methods. These methods are used to compare with ISO and ErrCor, as will be explained in the following two subsections promptly.

### *B. PSO Training Algorithm for tuning RBFN Parameters*

Particle Swarm Optimization (PSO) is a statistical optimization technique that is used to optimize the particles' parameters to find a global minimum for a specific population (swarm) problem [17], [34]–[36]. In RBFN literature [17], the swarm consists of particles representing all the RBFN parameters setting. The RBFN parameter setting includes all hidden units' parameters. Each $h$-hidden unit center, $\mathbf{c}_h$, width, $\sigma_h$, and weight, $w_h$, are included in each $j$-particle.

At time $t$, each $j$-particle has a position and a velocity, represented by vectors, $\mathbf{p}_j(t)$ and $\mathbf{v}_j(t)$, respectively:

$$\mathbf{p}_j(t) = \left[\, p_{j,1}(t), \dots, p_{j,m}(t), \dots, p_{j,M}\,(t)\right]^T \quad (7)$$

$$\mathbf{v}_j(t) = \left[\, v_{j,1}(t), \dots, v_{j,m}(t), \dots, v_{j,M}\,(t)\right]^T \quad (8)$$

wherein RBFN case, each $m$-element in particle $\mathbf{p}_j(t)$ in (7) represents the $h$-hidden unit parameters as follows:

$$p_{j,m}(t) = [\, \mathbf{c}_m^T, \sigma_m, w_m]^T \quad (9)$$

The bias is the last $M$ parameter and is considered a hidden unit with an only the weight parameter $p_{j,M}(t) = [\, w_0]^T$ where $M = H + 1$. Each $j$-particle keeps track of the best previous position and the best swarm position. The best previous position vector, $\mathbf{bp}_j(t)$, and the best overall swarm position vector, $\mathbf{Bp}(t)$, are represented respectively as:

$$\mathbf{bp}_j(t) = \left[bp_{j,1}(t), \dots, bp_{j,m}(t), \dots, bp_{j,M}\,(t)\right], \quad (10)$$

$$\mathbf{Bp}(t) = [Bp_1(t), \dots, Bp_m(t), \dots, Bp_M(t)]. \quad (11)$$

As demonstrated in Fig. 2, the new velocity at a time $(t+1)$ for $j$-particle is calculated as:

$$\mathbf{v}_j(t+1) = \omega\, \mathbf{v}_j(t) + \rho_1 r_1 \left(\mathbf{bp}_j\,(t) - \mathbf{p}_j(t)\right) + \rho_2 r_2 \left(\mathbf{Bp}_j(t) - \mathbf{p}_j(t)\right), \quad (12)$$

where $\omega$, is the inertia weight; $\rho_1$ and $\rho_2$ are constants, and $r_1$ and $r_2$ are random positive constants between (0, 1) which represent the cognitive and the social learning factors, respectively [37]. The new position, $\mathbf{p}_j(t+1)$, the new previous position, $\mathbf{bp}_j(t+1)$, and the new best position, $\mathbf{Bp}_j(t+1)$, are updated respectively as:

$$\mathbf{p}_j(t+1) = \mathbf{p}_j(t) + \mathbf{v}_j(t+1) \quad (13)$$

$$\mathbf{bp}_j(t+1) = \begin{cases} \mathbf{bp}_j(t), & if\ f\left(\mathbf{p}_j(t+1)\right) \geq f(\mathbf{bp}_j(t)) \\ \mathbf{p}_j(t+1), & otherwise \end{cases} \quad (14)$$

$$\mathbf{Bp}(t+1) = \arg\min_{\mathbf{bp}_j}\left(f\left(\mathbf{bp}_j(t+1)\right)\right), j = 1..J \quad (15)$$

where $f(.)$ is the fitness function used to minimize each entry of the particles. In the RBFN subject, the fitness function $f(.)$ represents the convergence cost function (3).

The PSO required a fixed number of swarm particles $J$ and a fixed number of particle parameters $M$, which are set intuitively by experiments. Each particle's position and velocity are randomly initiated and distributed over the searching space. The PSO algorithm repeatedly calculates particles' new positions, compares their fitness values, and directs these particles towards the least particle cost until convergence [38].

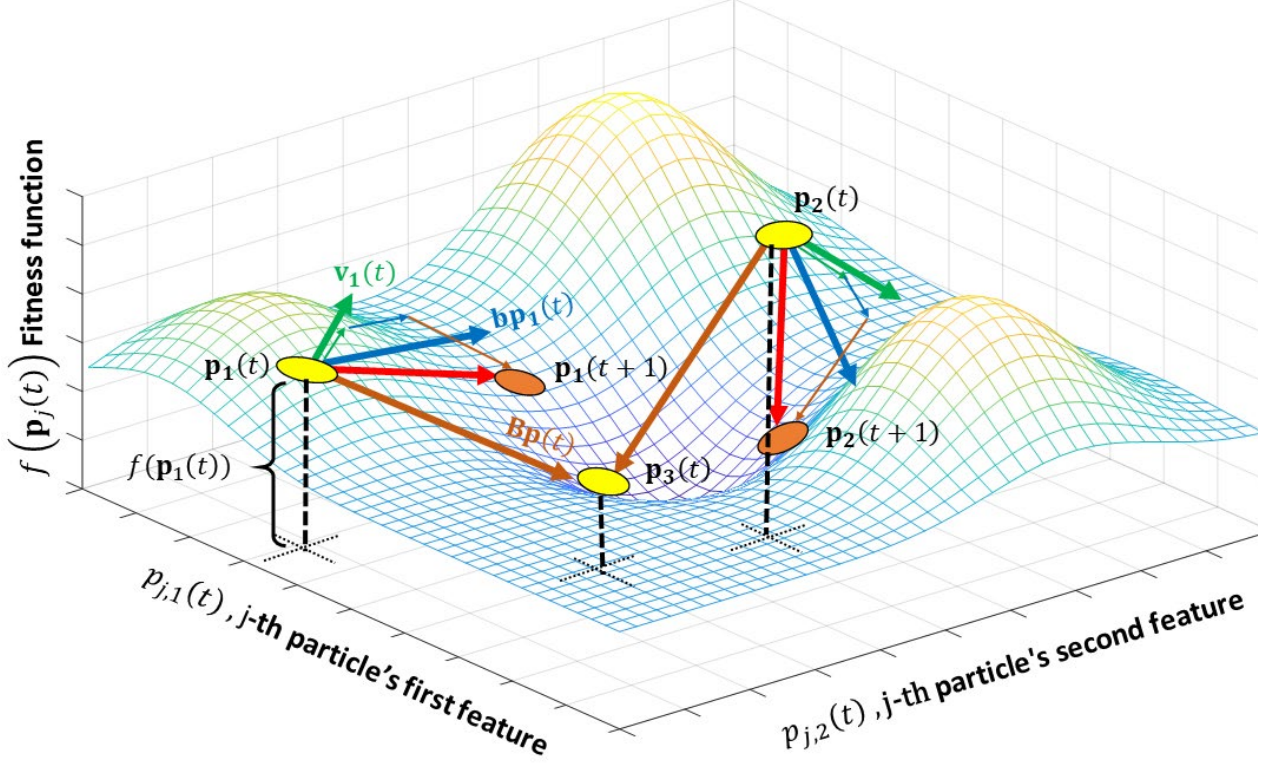

Fig. 2. An example of how the swarm particles of the PSO algorithm fly towards the best particle over the cost (fitness) function surface. $\mathbf{p}_1(t)$, $\mathbf{p}_2(t)$ and $\mathbf{p}_3(t)$ are three particle positions, at time $t$. $\mathbf{p}_3(t)$ has the less fitness, hence is has the best particle position in the swarm. PSO adjusts particles $\mathbf{p}_1(t)$ and $\mathbf{p}_2(t)$ positions based on their current best position $\mathbf{bp}_j(t)$, vecolity $\mathbf{v}_j(t)$ and best particle position $\mathbf{Bp}(t)$. $\mathbf{p}_1(t+1)$ and $\mathbf{p}_2(t+1)$, represent the new positions for particles $\mathbf{p}_1$ and $\mathbf{p}_2$, at time $t+1$, respectively.

### *C. APSO Training Algorithm for Tuning RBFN Parameters*

The APSO algorithm adjusts the particles' inertia weight and/or the number of hidden units while converging to an optimum solution [17]. One major problem in the RBFN training process is empirically setting the number of hidden units. However, increasing or pruning the number of hidden units in APSO tends to slow down the learning process. Adding a new hidden unit requires adding the unit parameters to each particle. Likewise, removing a unit requires removing its parameters from each swarm particle. This process tends to excessively use memory even with the suggestion of preserving a maximum empty space for the future expected neurons in the particle's memory size [17]. In our work, we consider only the inertia weights to adapt the particles' speed. The number of hidden units is set empirically to avoid extra delays in training.

The PSO heavily depends on the particles' inertia weight, which controls the flight of the swarm particles [37]. The inertia weight parameter is adaptively adjusted using *the diversity of particles*. The diversity of particles $D(t)$ at time $t$, is defined as the ratio of the minimum fitness value to the maximum fitness value of all particles, as follows [17]:

$$D(t) = \frac{\min_{f(\mathbf{p}_j(t))}\{x|x = f(\mathbf{p}_j(t)), \forall j = 1..J\}}{\max_{f(\mathbf{p}_j(t))}\{x|x = f(\mathbf{p}_j(t)), \forall j = 1..J\}} \quad (16)$$

A nonlinear *regressive function* is used to balance the effect of the diversity ratio:

$$\mathbf{H}(t) = \left(\alpha - D(t)\right)^{-1}, \quad (17)$$

where $\alpha \geq 2$ is a constant. The *fitness change ratio* of each $j$ −particle to the fitness of the best swarm particle is calculated as:

$$\mathbf{F}_j(t) = f(\mathbf{Bp}(t))/f(\mathbf{p}_j(t)). \quad (18)$$

The nonlinear regressive function (17) and the fitness change ratio (18) adjust each $j$-particle inertia weight, $\omega_j$, as:

$$\omega_j(t) = \mathbf{H}(t)\left(\mathbf{F}_j(t) + \beta\right), \quad (19)$$

where, $\beta \geq 0$, is a predefined positive constant. The APSO algorithm is briefly illustrated in Algorithm 1.

### *D. MC-PSO and MC-APSO Training Algorithms*

The MCRN is a faster, more accurate multi-column alternative to the single-structured RBFN using ISO and ErrCor training

Algorithm 1: APSO for training single-structured RBFN

```
1: Input: {x, y_d}= (training dataset), (RBFN : RBFN initial training model)
2: Output: =RBFN trained structure of RBFN.
3:  Initialize parameters:  ω(0), ρ1, ρ2, r1, r1, α, β ;
    All particles vectors: p(0), v(0), bp(0), Bp(0)
4:  for t = 1 to iterMax                          % iterations loop
5:      for j = 1 to J                            % particles loop
6:          Calculate fitness:   f(p_j(t))
7:          If fitness ≤ tolerance      %stopping convergence condition
8:              Set Model RBFN
9:              exit function
10:     end for
11:     Calculate particles diversity D(t)
12:     Calculate regressive function H(t)
13:     for j = 1 to J
14:         Calculate particles fitness change ratio F_j (t)
15:         Calculate particle inertia weight ω_j(t)
16:         Update particle vectors p_j(t + 1), v_j(t + 1), bp_j(t + 1),
17:     end for
18:     Update best swarm particle Bp(t + 1)
19: end for
```

algorithms [6]. MCRN, unlike ensemble learning, divides the dataset into $N$ subsets using the k-d tree algorithm. The $N$-chopped subsets are used to train $N$-individual RBFNs using ErrCor algorithm. The ErrCor increasingly inserts a single unit with each training epoch until convergence. Each trained individual RBFN becomes specialized NNs on their subset regions. This paper uses the same MCRN process for dividing the dataset into subsets using the k-d tree algorithm with our two novel MC-PSO and MC-APSO structures. Each individual RBFN is trained using either PSO or APSO, as will be explained promptly.

Let $S$, represents the set of all input vectors of the dataset, and $s_n$ is the $n$-th chopped subset from that dataset, then the k-d tree chopping process produces $N$ subset as follows:

$$S = s_1 \cup s_2 \ldots \cup s_n \ldots \cup s_N \tag{20}$$

Each $s_n$ is used to train $n$-th RBFN using either PSO or APSO. The PSO and APSO explained in subsections III.B and III.C, respectively, are used to tune the local parameters for each RBFN. Each training algorithm uses a fixed number of hidden units and a fixed number of particles. The initial values of these particles' positions $\mathbf{p}_j^n(t)$, and their velocity vectors $\mathbf{v}_j^n(t)$, are randomly initialized. The algorithm keeps track of the best particle position $\mathbf{bp}_j^n(t)$ and best swarm position $\mathbf{Bp}^n(t)$, where the super suffix $n$ represents the $n$-th subset used to train the $n$-th RBFN.

The PSO updates the swarm parameters accordingly and (13) - (15) will be as follows:

$$\mathbf{p}_j^n(t+1) = \mathbf{p}_j^n(t) + \mathbf{v}_j^n(t+1) \tag{21}$$

$$\mathbf{bp}_j^n(t+1) = \begin{cases} \mathbf{bp}_j^n(t), & if\ f\left(\mathbf{p}_j^n(t+1)\right) \geq f(\mathbf{bp}_j^n(t)) \\ \mathbf{p}_j^n(t+1), & otherwise \end{cases} \tag{22}$$

$$\mathbf{Bp}^n(t+1) = \arg\min_{\mathbf{bp}_j}\left(f\left(\mathbf{bp}_j^n(t+1)\right)\right), \quad j = 1..J, n = 1..N \tag{23}$$

The PSO continues to update each individual RBFN parameters until the fitness function in (3) converges to a tolerance condition.

The APSO training algorithm updates the inertia weight in addition to the swarm particle parameters. Therefore, the $n$-th subset diversity of particles in (16), $D^n(t)$, at time $t$ is:

$$D^n(t) = \frac{\min_{f(\mathbf{p}_j^n(t))}\{x|x=f(\mathbf{p}_j^n(t)), \forall j=1..J\}}{\max_{f(\mathbf{p}_j^n(t))}\{x|x=f(\mathbf{p}_j^n(t)), \forall j=1..J\}}. \tag{24}$$

The nonlinear regressive function for the $n$-th subset in (17) is set to:

$$\mathbf{H}^n(t) = \left(\alpha - D^n(t)\right)^{-1}. \tag{25}$$

Likewise, the $n$-th subset fitness change ratio of each $j$ −particle to the fitness of the best swarm particle in (18), is calculated as:

$$\mathbf{F}_j^n(t) = f(\mathbf{Bp}^n(t))/f(\mathbf{p}_j^n(t)). \tag{26}$$

Finally, the $j$-particle inertia weight of the $n$-th subset for, $\omega_j^n$, in (19) is calculated as follows:

$$\omega_j^n(t) = \mathbf{H}^n(t)\left(\mathbf{F}_j^n(t) + \beta\right), \tag{27}$$

During runtime, the trained individual RBFNs are stacked in a multi-column structure as shown in Fig. 3. Once a new testing data vector $\tilde{\mathbf{x}}$ is applied, the ***Subset Selector*** uses a KNN technique to find the $k$ nearest neighbor's vectors from the training

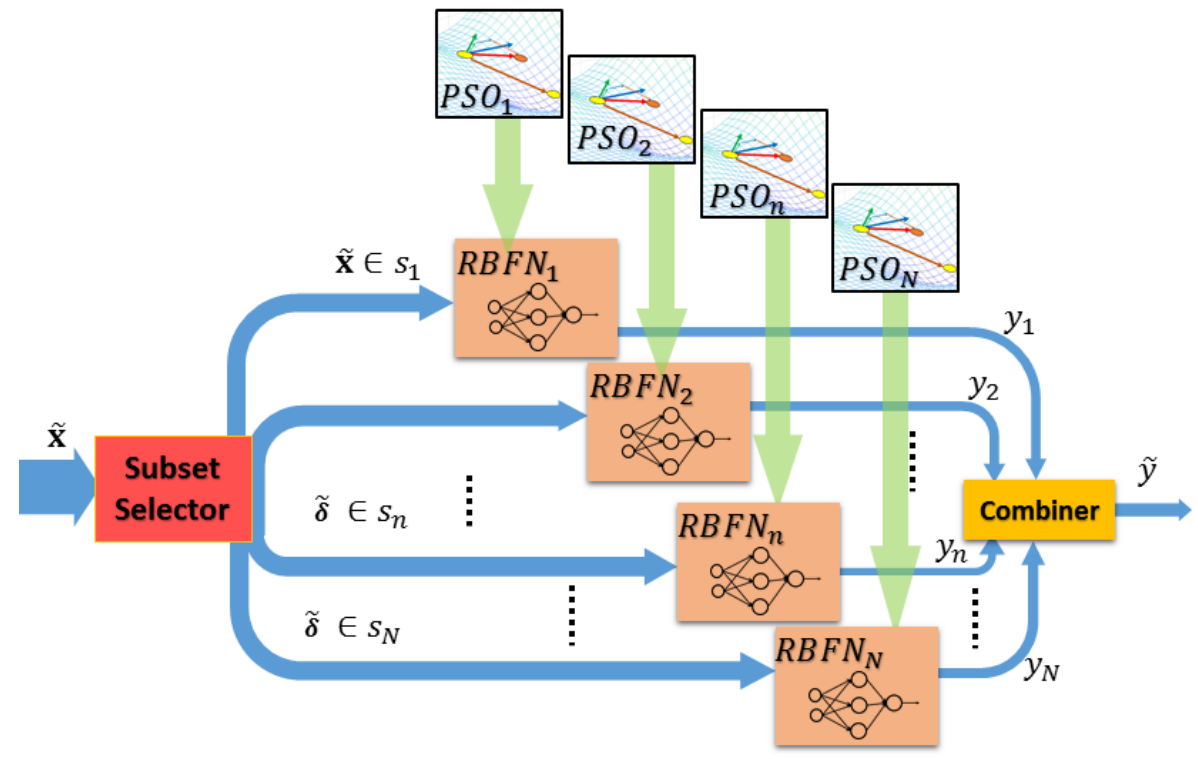

Fig. 3. Internal structure of the MC-PSO and the MC-APSO (replacing PSO by APSO) with the input subset selector, $N$ individual RBFNs, the output combiner, and PSO.

instances to that input vector [6]. The $k$ value is a predefined odd integer number, and it is odd to break the tie in comparable cases. These $k$ nearest neighbor vectors were previously used in training $k$ RBFNs (dubbed as ***neighbor RBFNs***). The input vector $\tilde{\mathbf{x}}$ is forwarded to those neighbor RBFNs, and each one gives its output $\tilde{y}_k$ to the ***output combiner***. The latter calculates the single multi-column structured system output $y$. This process is more explained in the following steps:

1- The ***Subset Selector*** determines the Euclidean distance between the new testing input vector, $\tilde{\mathbf{x}}$, and each training vector, $\mathbf{x}^p$. KNN chooses $k$ nearest instances with the minimum Euclidean distances $d_k$, as in:

$$\mathbf{d}_k = \min_{1\ldots k}(\|\tilde{\mathbf{x}} - \mathbf{x}^p\|), \quad p = 1 \ldots P, \qquad (28)$$

where, $\mathbf{d}_k = \{d_1, d_2, \ldots, d_k\}$; and $d_k$ is the $k$-th minimum Euclidean distance between the new testing point $\tilde{\mathbf{x}}$, and the $k$-th minimal training point. These $k$ nearest vectors belong to $l$ subsets $\delta_l$; where $l \leq k$. Some neighbor vectors may belong to the same subset as shown in the graphical example in Fig. 4. Only $l$ NNs are selected to be executed and these NNs are the ones trained by the $l$ selected subsets as follows:

$$\delta_l = \{\forall s_n, s_n \subset S, \mathbf{x}^k = \arg_{\mathbf{x}^p} \min_{1\ldots k}\|\tilde{\mathbf{x}} - \mathbf{x}^p\| \in s_n\} \quad (29)$$

where $\delta_l$ is the $l$-th selected subsets from all $S$ subsets.

2- The ***Individual NNs*** which represent the $l$ selected RBFNs, work individually in parallel to produce $l$ results $\tilde{y}_l$ for a test input $\tilde{\mathbf{x}}$, as follows:

$$\tilde{y}_l = \sum_{h=0}^{H} w_h^l \theta_h^l(\tilde{\mathbf{x}}), \forall \delta_l, \qquad (30)$$

where $\tilde{y}_l$ represents the output decision of the $l$ selected NN given input $\tilde{\mathbf{x}}$.

3- The ***Combiner*** collects only $l$ NNs results while the remaining $N - l$ RBFNs are not used (rested) for the current input test vector $\tilde{\mathbf{x}}$. The outputs of $l$ NNs, $\tilde{y}_l$, contribute to the overall output decision $y$. In the case of a subset that has more than one neighbor, the individual RBFN of that subset will fire ones but the output results are multiplied by the number of included neighbor instances. The overall MCRN output $y$ is calculated based on the average sum of all $k$ outputs as, $\tilde{y} = 1/k \sum_{i=1}^{k} \tilde{y}_i$. In classification problems, the real value output $y$ should be hard limited to give the duple class values.

The best-case scenario is when some of or all the $k$ neighbors belong to the same subset(s). In this case, the selected $l$-NNs are less than $k$, while the NN, which has many neighbors, runs only once, and more NNs are not used (rested) during testing. The test input vector is, confidently, more related to that duplicated subset; hence, the decision of those NNs is more weighted toward guessing the correct output. The training and testing of the MC-APSO algorithm are shown in more detail in Algorithm 2.

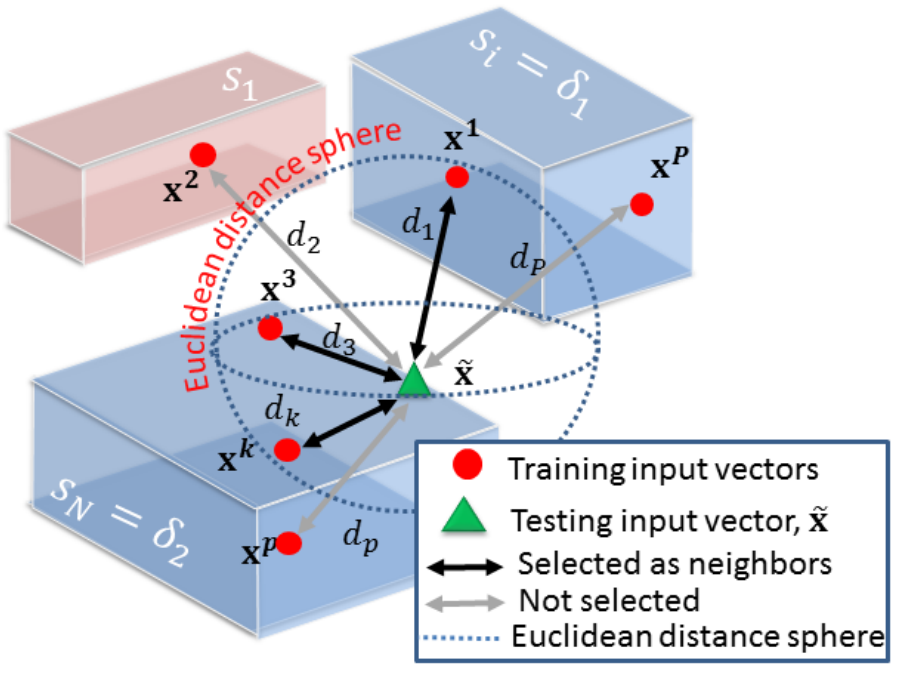

Fig. 4. 3D illustration of the KNN algorithm that selects $k = 3$ neighbor vectors. Each selected vector is located at a subset $s_i \subset S$. It also shows the Euclidean distance sphere where the neighbors are located inside it. We can also note that vector $\mathbf{x}^1$ belongs to subset $s_i = \delta_1$, whereas, vectors $\mathbf{x}^3$ and $\mathbf{x}^k$ are both neighbors and also belongs to same subset $s_N = \delta_2$.

ALGORITHM 2: TRAINING AND TESTING THE MULTI-STRUCTURED MC-APSO

*Training:*
1: Input: $\{\mathbf{x}, y_d\}$= (training dataset), ($RBFN$ : RBFN initial training model)
2: Output: =MC-APSO trained structure with $N$ RBFNs.
3: for each subset $n = 1\ to\ N$, ------ **in parallel**
4: Initialize parameters: $\omega^n(0), \rho_1, \rho_2, r_1, r_1, \alpha, \beta$ ;
5: Initialize $n$ RBFN with random numbers.
6: Initialize each swarm $j$ −particle parameters $\mathbf{p}_j^n, \mathbf{v}_j^n, \mathbf{bp}_j^n$ and $\mathbf{Bp}^n$
7: Train subset $n$ with APSO as in Algorithm 1
8: end for

*Testing:*
1: Input: $\tilde{\mathbf{x}}$, (MC-APSO : trained model)
2: Output: $\tilde{y}$
3: Select $k$ nearest neighbors from all dataset
4: Select the $\delta_l$ subset that has the k neighbors
5: for each $j = 1$ to $l$ subset, -------**in parallel**
6: Calculate $\tilde{y}_l$ by running the $l$ RBFNs
7: end for
8: Combine and average the $k$ outputs $\tilde{y} = 1/k\ \sum_{i=1}^{k} \tilde{y}_i$

In our work, we divide each dataset into two separate subsets. These subsets are experimentally trained using RBFNs structure with either PSO or APSO algorithms. The RBFNs are stacked in parallel during the testing phase. The results are compared to single-structured RBFNs results with different gradient and population training algorithms, as will be explained in detail in the following experimental section.

## IV. EXPERIMENTAL EVALUATION

This section is organized as follows: In subsection A, a variety of generic benchmark datasets and their characteristics are explained, in addition to the setting of all research experiments. Subsection B presents results of training a single-structured RBFN using traditional training algorithm, including gradient algorithms, such as ISO and ErrCor, and population algorithms, such as PSO and APSO. Subsection C shows the experimental results of the two proposed multi-column RBFNs training methods, the MC-PSO and the MC-APSO. Subsection D compares the training and testing speed of all previously listed algorithms in subsections IV.B, and IV.C. General Comparisons of the MC-PSO and MC-APSO methods to evolutionary and MCRN methods are discussed in last subsection.

### *A. Dataset Characteristics and Experimental Setting*

This paper uses 24 benchmark UCI datasets [39] to test, compare, and evaluate each experimental RBFN training algorithm (Table I). These alphabetically ordered datasets show variety in the number of features, number of classes, number of instances, and variety of *class density* values. The *class density* represents the ratio of the number of positive class instances to the overall number of instances, as shown in Table I. The diversity in parameters and densities provides different challenges in training an RBFN. Training a specific dataset using a particular algorithm varies and can be difficult from one algorithm to another. The key factors for a successful algorithm are accurate results, fast training convergence, and rapid response to an input test vector.

TABLE I
DATASET PARAMETERS

| Dataset | Features | Classes | Instances | Density (%) |
|---|---|---|---|---|
| Banknote | 4 | 2 | 1372 | 44.46 |
| Dermatology | 33 | 6 | 366 | 30.6, 16.67, 19.67, 13.39, 14.21, 5.46 |
| Ecoli | 7 | 8* | 336 | 42.56 |
| Glass-Id | 9 | 2 | 214 | 23.83 |
| Haberman | 3 | 2 | 306 | 73.53 |
| Heart Cleveland | 13 | 5* | 297 | 53.87 |
| Hepatitis | 19 | 2 | 155 | 20.65 |
| Ionosphere | 34 | 2 | 351 | 64.1 |
| Iris | 4 | 3 | 150 | 33.3, 33.3, 33.3 |
| Libras Mov. | 90 | 15* | 360 | 6.67 |
| Liver | 5 | 2 | 341 | 41.64 |
| Magic G. Telescope | 10 | 2 | 19020 | 64.84 |
| Musk | 166 | 2 | 476 | 43.49 |
| Parkinsons | 22 | 2 | 195 | 75.38 |
| Pima India | 8 | 2 | 768 | 34.9 |
| Seeds | 7 | 3 | 210 | 33.3, 33.3, 33.3 |
| Segment | 19 | 7* | 2310 | 14.29 |
| Sonar | 60 | 2 | 208 | 53.37 |
| Thyroid | 21 | 3 | 7200 | 2.3, 5.1, 92.6 |
| Transfusion | 4 | 2 | 748 | 23.8 |
| Vertebral | 6 | 3 | 310 | 19.35, 32.26, 48.39 |
| Vowel | 10 | 11* | 990 | 9.09 |
| Wisconsin | 9 | 2 | 699 | 65.52 |
| Yeast | 8 | 10* | 1484 | 31.2 |

*The first output is used in these datasets and the reported density is the density of that output only.

For statistically assessing, each experiment performance in this research, a stratified five-fold cross-validation technique is conducted using Leave-one-out cross-validation technique. With each experimental training round, four folds are used interchangeably for training and one for testing. Some datasets require normalization to set all features into comparable values. For the datasets that have multiple classes, a one-vs-all technique is used to obtain the desired output in all training and testing results. A single output is used for some datasets with multiple classes for easy comparison. All results were obtained using a Dell 7910 series workstation, with 128GB RAM, dual 3.5 GHz, 15M L1 Cache, Xeon processor, and a dual NVIDIA 24GB VRAM. The operating system is Windows 10, using the multi-core MATLAB 2018 software package. The multi-core feature is used to perform parallelism in the proposed multi-column techniques.

### *B. Training RBFN Using Different Training Techniques:*

In this subsection, accuracy, and recall comparisons are shown for the results of training a single-structured RBFN using four different state-of-the-art RBFN gradient and population training techniques. The ISO [7] and ErrCor [9], previously discussed in subsection III.A are two gradient methods where the parameters are tuned using gradient-descent optimization methods. These techniques are compared to their counterparts, the particle swarm population methods; the PSO and the APSO, which are previously explained in subsections III.B and III.C, respectively.

For comparison purposes, we limited every single-structured RBFN to have the same number of maximum hidden units. Each dataset in Table I is trained and tested using all four-training algorithms. Table II shows the accuracy and recall results of using ISO, ErrCor, PSO, and APSO, respectively. The number of hidden units is set to $H = 9$ for each experiment. The number of swarm particles $J$ for the PSO and APSO algorithm in each experiment is also fixed, as shown in Table II. For all experiments, the PSO parameters are set as $\omega = 0.8,\ \rho_1 = 2, \rho_2 = 2, r_1 = 0.6$ and $r_1 = 0.3$. The inertia weight $\omega(t)$ for APSO algorithm is adaptively changing while the new parameters of APSO are set as $\alpha = 2$ and $\beta = 0.1$. Those values are suggested within suitable ranges by [17] and [16]. We fixed those PSO and APSO hyperparameters in all our implementations for a fair comparison with our two novel MC-PSO and MC-APSO algorithms. Investigating the effect of changing those parameters is left for future development.

The accuracy and recall results in Table II show that the ISO and ErrCor outperform PSO and APSO in most cases, except for minor cases when PSO and APSO results are slightly better than ISO and ErrCor. However, the PSO and APSO results are still comparable to the best ones. ISO and ErrCor use the gradient of the feedback error to tune the RBFN's parameters. This technique ensures that the cost function is stepping confidently to an optimal solution until a convergence condition is met. On the other hand, PSO and APSO use flying particles and cognitive contributions to find a global minimum in the cost function. The methods use a trade-off between the computational communications between those particles and each particle's decision to locate the most profound solution among them and try to converge toward the best particle position. While each particle has the whole structure of the RBFN parameters, these methods use a fitness function to optimize the algorithm, as explained previously in subsections III.B and III.C.

The ISO initiates the hidden units with random vectors taken from the training dataset with the same number of hidden units. Afterward, the ISO algorithm uses an improved second-order gradient algorithm to tune these hidden units until convergence. On the other hand, the ErrCor has an incremental RBFN structure. With each epoch, the algorithm selects a single vector from the training set and adds it as a new hidden unit. ErrCor, then, tunes the whole structure using the steepest gradients. The selected vector is the one that shows a maximum error violation (maximum training error) using the current RBFN structure.

TABLE II
ACCURACY AND RECALL RESULTS OF TESTING SINGLE-STRUCTURED RBFNS USING ISO, ERRCOR, PSO AND APSO ALGORITHMS.

| Dataset | $H/J$ | Accuracy | | | | Recall | | | |
|---|---|---|---|---|---|---|---|---|---|
| | | ISO | ErrCor | PSO | APSO | ISO | ErrCor | PSO | APSO |
| Banknote | 9/35 | 90.89±5.37 | **100.00±0.00** | 98.98±0.83 | 95.70±6.41 | 94.76±4.49 | **100.00±0.00** | 97.78±1.78 | 92.27±10.36 |
| Dermatology | 9/25 | **99.18±0.60** | 98.54±0.83 | 98.90±1.07 | 95.44±3.74 | **97.53±1.79** | 95.61±2.48 | 96.70±3.20 | 86.33±11.23 |
| Ecoli | 9/25 | 94.64±1.69 | 95.53±1.84 | **96.43±2.49** | 96.13±2.26 | **96.32±2.53** | 95.81±3.77 | 95.22±3.90 | 95.16±3.11 |
| Glass-Id | 9/15 | **92.51±3.50** | 91.06±4.01 | 91.07±3.90 | 92.49±3.57 | 85.08±14.68 | **87.78±8.24** | 87.62±13.64 | 84.73±11.94 |
| Haberman | 9/25 | **75.81±2.54** | 72.88±2.65 | 74.84±1.41 | 75.48±2.09 | **79.77±3.08** | 78.42±1.69 | 76.11±1.63 | 77.40±1.54 |
| Heart Cleveland | 9/35 | 76.76±4.09 | 82.47±2.94 | 82.47±3.60 | **83.15±2.48** | **86.43±4.29** | 83.62±2.91 | 79.78±3.74 | 81.31±2.76 |
| Hepatitis | 9/25 | **86.43±4.27** | 79.47±10.07 | 81.26±4.38 | 80.70±8.65 | **81.00±20.74** | 58.79±26.96 | 60.57±25.73 | 64.29±29.06 |
| Ionosphere | 9/39 | **92.88±0.97** | 88.32±5.28 | 86.62±4.99 | 88.89±3.83 | **95.99±2.61** | 91.16±4.26 | 87.42±3.22 | 88.51±3.55 |
| Iris | 9/15 | 97.33±2.43 | **97.78±2.72** | 96.00±1.86 | 94.67±2.53 | 96.00±3.65 | **96.67±4.08** | 94.00±2.79 | 92.00±3.80 |
| Libras Mov. | 9/35 | **97.21±1.41** | 96.10±1.20 | 95.83±1.42 | 95.00±1.57 | 95.00±11.18 | 84.29±22.81 | 90.00±22.36 | **100.00±0.00** |
| Liver | 9/35 | 65.40±6.42 | **69.80±7.32** | 66.59±6.88 | 68.93±6.42 | 63.44±12.46 | **68.08±10.66** | 60.79±8.47 | 65.10±8.00 |
| Magic G. Telescope | 9/35 | 74.31±3.57 | **84.99±3.23** | 81.05±1.67 | 82.23±0.76 | 78.33±4.44 | **84.96±2.34** | 81.66±1.58 | 82.26±0.60 |
| Musk | 9/33 | 86.98±2.18 | **90.13±2.01** | 74.15±6.79 | 71.43±4.02 | **88.85±5.58** | 87.15±3.28 | 74.02±11.23 | 69.15±6.30 |
| Parkinsons | 9/27 | 84.61±7.91 | **87.68±8.00** | 83.60±3.83 | 80.49±3.05 | 84.64±7.15 | **89.30±9.85** | 83.33±3.14 | 83.77±4.08 |
| Pima India. | 9/29 | 70.31±0.95 | **75.00±2.88** | 73.56±5.11 | 74.74±3.75 | 67.80±7.44 | **70.01±8.15** | 63.94±7.84 | 69.26±10.01 |
| Seeds | 9/15 | **95.56±3.05** | 93.65±2.97 | 94.60±1.81 | 94.60±1.81 | **93.33±4.58** | 90.48±4.45 | 91.90±2.71 | 91.90±2.71 |
| Segment | 9/35 | 98.48±0.59 | **99.70±0.25** | 98.79±0.39 | 98.14±1.67 | 99.02±1.47 | **99.69±0.69** | 99.35±1.44 | 99.26±1.03 |
| Sonar | 9/29 | **85.10±2.01** | 77.89±1.85 | 74.51±7.04 | 75.42±6.07 | **87.51±5.34** | 81.60±8.04 | 74.69±10.11 | 76.35±8.74 |
| Thyroid | 9/35 | 95.48±0.36 | 96.22±0.34 | 95.69±0.08 | **96.29±0.28** | 93.22±0.55 | 94.33±0.51 | 93.53±0.12 | **94.43±0.41** |
| Transfusion | 9/29 | **79.95±2.49** | 78.48±2.85 | 78.75±1.98 | 79.55±3.19 | **65.22±9.21** | 58.70±10.59 | 59.42±6.93 | 61.45±13.18 |
| Vertebral | 9/35 | 87.96±2.98 | 89.25±2.01 | **89.68±3.37** | 86.45±3.10 | 81.94±4.48 | 83.87±3.02 | **84.52±5.05** | 79.68±4.65 |
| Vowel | 9/29 | 98.48±0.51 | **99.09±0.90** | 96.57±1.31 | 95.76±1.36 | **100.00±0.00** | 97.61±3.30 | 93.97±8.45 | 89.30±7.28 |
| Wisconsin | 9/25 | 96.15±2.33 | 96.00±2.10 | **97.29±1.61** | 97.00±1.26 | 97.40±2.30 | 97.39±2.31 | **98.88±0.79** | 98.66±0.51 |
| Yeast | 9/25 | **72.64±1.70** | 72.10±2.92 | 72.24±3.04 | 70.49±2.89 | **61.89±6.26** | 57.87±7.12 | 59.27±7.14 | 54.15±6.96 |

The advantage of using PSO and APSO over the gradient-descent methods is the fast training convergence to an optimum solution, as will be shown later in this research. PSO and APSO do not have complex gradient computations and no excessive use of kernel inner products. A straightforward fitness function is a comparing judge in converging to a fine-tuned NN.

### C. *MC-PSO and MC-APSO Training Techniques*

The PSO and APSO show promising and comparable accuracy and recall results to those of the ISO and ErrCor in the previous subsection. In this subsection, we demonstrate and discuss the accuracy and recall results of the proposed multi-column methods of subsection III.D and compare them to the results of all four single-structured results shown in subsection IV.B.

The MCRN, as a multi-column structured method, showed improvement to the ErrCor method both in terms of results and speed perspectives [6]. Dividing the dataset into smaller subsets decreases the number of instances per training pattern hence the training computations for each subset. Each subset is used to train an individual RBFN in parallel, which becomes a specialized trained NN on its local subset region. When testing is conducted, those Individual RBFNs are also run in parallel and give their output desired suggestions to the output combiner. The selected outputs are weighted-summed and averaged to produce the overall multi-column output.

The experiments in Table III demonstrate the accuracy and recall results of the two proposed multi-column structured results: the MC-PSO and the MC-APSO. The MC-PSO uses the PSO as the training algorithm of the individual RBFNs, while MC-APSO uses APSO, as previously explained in subsection III.C. Each dataset is divided into two subsets, and each subset is used separately to train an individual RBFN with a fixed number of hidden units, $H$, and a fixed number of swarm particles, $J$. The number of hidden units is intentionally limited not to exceed $H \leq 9$ to compare with those in Table II.

During testing time, the new input entry passes through a KNN algorithm, which determines the $k$ neighbors to the input. Each training instance is tagged with the subset that it belongs to. The $k$ instance neighbors indicate the individual RBFNs that used to train their subsets. Each selected individual RBFN has $H$ hidden units and is fed with that new test input entry. The selected RBFNs will pass their suggested output to the output combiner; where the weighted sum of those outputs is hard limited to produce the output of the multi-column structure.

To conduct comparability, each experiment in Table III is also performed with the same five-fold cross-validation technique used in Table II. Each experimental dataset is chopped through the $h$ feature using the k-d tree algorithm. Although the density and the distribution of the training instances for the other four folds are changing with each cross-validation cycle, the results are reported using the same $h-$chopping dimension and the same number of $k$ neighbors for fair comparisons with other algorithms. These results are reported in Table III as the numbers without parentheses (we named them as *non-preferable selections*). Nevertheless, it is beneficial to show the results of selecting the best chopping results regardless of keeping the same chopped dimension or the same number of neighbors. These results are reported in Table III as the numbers between parentheses (we named them as *preferable selections*).

The non-preferable selection results in Table III show superior improvements in most dataset experiments in accuracy and recall over the four traditional single-structured algorithms shown in Table II. For example, the maximum accuracy improvements of MC-PSO reach 8.45%, 4.43%, 3.63%, and 6.74% over ISO in Banknote, ErrCor in Hepatitis, PSO in Parkinsons, and APSO in Parkinsons, respectively. Similarly, MC-APSO showed maximum accuracy improvements of reach 8.53%, 6.31%, 5.31%, and 7.71% over ISO in Magic G. Telescope, ErrCor in Hepatitis, PSO in Sonar, and APSO in Parkinsons, respectively. In most cases, the MC-PSO and MC-APSO significantly outperform their antecedent algorithms: the PSO and APSO. Needless to mention that the preferable selection results tip the scale and improve outcomes even better.

TABLE III
ACCURACY AND RECALL RESULTS OF TESTING MULTI-COLUMN RBFN USING PSO AND APSO ALGORITHMS. RESULTS SHOWS ***NON-PREFERABLE*** SELECTION OF CROSS-VALIDATED CHOPPING FEATURES (WITHOUT PARENTHESES) AND ***PREFERABLE*** ONES (WITH PARENTHESES)

| Chopped Dataset | PSO setting | | APSO setting | | Accuracy | | Recall | |
|---|---|---|---|---|---|---|---|---|
| | $H/J$ | $h/k$ | $H/J$ | $h/k$ | MC-PSO | MC-APSO | MC-PSO | MC-APSO |
| Banknote | 9/25 | 4/7 | 9/25 | 1/1 | **99.34±0.87** (99.85±0.2) | 98.69±1.14 (99.49±0.55) | 98.57±1.86 (99.67±0.45) | **99.01±0.35** (100.00±0.0) |
| Dermatology | 9/19 | 4/3 | 9/25 | 18/7 | **98.13±0.80** (98.49±0.75) | 96.40±1.30 (97.31±0.89) | 96.87±2.29 (100±0.00) | **97.28±2.66** (98.38±2.84) |
| Ecoli | 5/15 | 2/5 | 5/15 | 3/5 | 96.72±3.24 (98.51±1.83) | **97.32±2.44** (97.62±2.26) | 94.54±3.23 (97.95±3.07) | **95.31±3.79** (97.13±3.07) |
| Glass-Id | 7/19 | 4/7 | 7/19 | 4/7 | **93.95±4.80 (**96.72±1.30) | 93.91±2.67 (95.77±2.03) | 87.70±4.00 (96.18±5.24) | **91.94±7.63** (97.78±4.97) |
| Haberman | 7/19 | 1/7 | 9/19 | 1/1 | 76.13±2.87 (77.44±2.24) | **77.12±2.32** (78.43±1.40) | 77.17±1.73 (78.46±1.90) | **79.13±4.06** (80.10±3.90) |
| Heart Cleveland | 7/25 | 13/7 | 7/25 | 9/5 | 83.19±2.99 (86.87±2.19) | **83.82±3.79** (86.85±4.40) | 82.10±5.53 (87.16±5.89) | **83.12±4.60** (86.84±3.73) |
| Hepatitis | 5/25 | 16/3 | 9/25 | 2/7 | 83.90±6.69 (89.03±2.90) | **85.78±4.94** (89.69±2.65) | **71.90±27.06** (91.00±12.45) | 70.17±24.62 (95.00±11.18) |
| Ionosphere | 5/25 | 20/7 | 7/25 | 14/3 | 88.33±3.88 (91.18±3.36) | **89.19±4.50** (92.89±3.60) | **88.28±5.27** (94.91±2.31) | 86.74±3.99 (94.84±3.07) |
| Iris | 5/12 | 3/3 | 5/12 | 4/3 | **97.11±2.56** (97.56±2.14) | 94.22±2.53 (95.11±2.56) | **95.98±3.65** (97.95±3.07) | 92.14±5.73 (95.71±3.92) |
| Libras Mov. | 7/25 | 90/7 | 7/25 | 35/3 | **96.66±1.26** (98.06±0.76) | 96.39±0.74 (98.61±0.98) | 90.00±14.91 (100.00±0.00) | **95.00±11.18** (100.00±0.00) |
| Liver | 5/15 | 2/7 | 5/25 | 6/7 | 69.52±6.73 (71.87±6.27) | **70.39±6.05** (72.74±2.73) | **71.82±10.40** (83.29±10.31) | 70.84±10.18 (76.79±5.92) |
| Magic G. Telescope | 9/25 | 5/7 | 9/25 | 5/7 | 81.18±0.57 (81.97±0.19) | **82.84±0.58** (83.44±0.28) | 81.57±0.36 (84.12±0.28) | **82.61±0.83** (84.31±0.47) |
| Musk | 9/33 | 161/5 | 9/33 | 27/7 | 76.28±7.25 (82.78±2.97) | **76.72±5.90** (83.42±4.01) | 78.26±8.65 (94.38±5.28) | **82.13±9.49** (92.45±6.38) |
| Parkinsons | 9/19 | 12/5 | 9/19 | 12/7 | 87.23±5.93 (87.23±5.93) | **88.20±6.69** (90.77±3.44) | 86.50±5.33 (86.50±5.33) | **88.24±4.16** (93.27±3.76) |
| Pima India. | 9/29 | 7/5 | 9/29 | 4/5 | 75.26±4.92 (78.38±3.23) | **76.82±2.17** (79.29±3.11) | **71.77±11.16** (79.62±8.98) | 70.50±4.78 (78.17±7.92) |
| Seeds | 5/25 | 3/1 | 7/25 | 3/7 | **95.56±2.07** (95.87±2.27) | 94.13±2.29 (95.56±2.61) | **93.94±4.81** (96.86±3.49) | 91.54±4.84 (95.97±3.73) |
| Segment | 9/25 | 11/5 | 9/25 | 9/7 | **98.83±0.83** (99.05±0.58) | 98.61±0.85 (99.18±0.49) | **100.00±0.00** (100.00±0.00) | 98.35±2.01 (100.00±0.00) |
| Sonar | 9/19 | 16/7 | 9/19 | 22/3 | **77.94±5.17** (87.01±2.80) | 79.82±4.66 (87.95±4.29) | 80.09±6.15 (92.22±4.97) | **81.78±7.13** (90.79±6.66) |
| Thyroid | 9/25 | 18/5 | 9/25 | 8/7 | 95.96±0.26 (96.19±0.19) | **96.29±0.35** (96.51±0.19) | 93.98±0.28 (94.52±0.18) | **94.42±0.52** (94.83±0.27) |
| Transfusion | 9/17 | 1/1 | 7/17 | 1/1 | **80.22±3.12** (81.29±2.29) | 80.22±2.40 (81.29±3.13) | **64.61±12.26** (68.08±7.84) | 64.34±8.05 (67.94±10.81) |
| Vertebral | 7/25 | 3/5 | 7/25 | 5/5 | **89.25±3.65** (90.22±2.80) | 88.49±2.42 (89.57±2.86) | **87.25±6.14** (89.94±3.98) | 85.18±4.35 (87.66±3.65) |
| Vowel | 9/19 | 10/1 | 9/29 | 10/1 | **97.58±0.83** (98.08±0.97) | 96.87±0.66 (97.47±0.36) | 96.24±5.29 (100±0.00) | **97.24±3.79** (100±0.00) |
| Wisconsin | 5/15 | 9/5 | 5/15 | 9/3 | 97.00±1.53 (97.43±1.39) | **97.43±1.70** (97.82±1.54) | 98.46±0.96 (98.68±1.19) | **99.11±0.92** (99.34±0.99) |
| Yeast | 5/10 | 8/7 | 9/19 | 7/5 | 72.24±2.25 (73.45±1.93) | **73.05±2.41** (74.40±1.69) | 61.92±7.74 (67.11±4.73) | **62.59±6.05** (66.07±3.21) |

Fig. 5 shows more comprehensive bar plots of percentage differences for the MC-PSO and the MC-APSO non-preferable results over the four single-structured traditional results. Fig. 5 summarizes all datasets' experiment accuracy and recall results shown in Table II and Table III in four percentage differences bar plots. The accuracy results in Fig. 5 (a) and (b) show winning cases of 18/24 and 16/24 for the MC-PSO and MC-APSO, respectively, over most of the other algorithms. On the other hand, the winning examples of the recall results are 15/24 and 19/24 for the MC-PSO and MC-APSO, respectively. Note that winning is achieved if the summation of the four results of ISO, ErrCor, PSO, and APSO tends to have positive value, while fail is when average is negative.

ErrCor and ISO show better results in some cases over MC-PSO and MC-APSO, as in Musk and Dermatology datasets. Where Musk dataset has the highest number of features among all datasets while the Dermatology dataset has an insufficient number of instances compared to the number of instances. For such conditions, population algorithms suffer from finding a global minimum while gradient descent methods are more likely to perform better. However, the MC-PSO and MC-APSO results in the Musk dataset experiments still beat their peer algorithms: the PSO and APSO, while Dermatology dataset suffers more complications when splitting the dataset with even less sufficient number of instances with MC-PSO but recovered with MC-APSO due to the later adaptive nature.

Generally, the MC-APSO leads to better results than the MC-PSO's in terms of accuracy and recall, except for a dataset with a small number of instances. The advantage of the adaptive PSO training for the individual RBFNs in the MC-APSO tips the scale over MC-PSO, which was trained with traditional PSO. The bolded font in Table III shows winning cases with accuracy results of 13/24 of MC-APSO over MC-PSO, while the recall results for winning cases are 15/24. Even with the remaining losses, the MC-PSO leads with a small percentage. A good advantage is that the new method improved the results of both PSO and APSO shown in Table II. Similar to the training of PSO vs. APSO, MC-PSO is still faster than MC-APSO, as follows.

### *D. Speed Comparison of Single-Structured to MC-PSO and MC-APSO Techniques*

Other important factors in evaluating a training algorithm and an NN structure are the convergence speed of that algorithm to meet an optimal solution and the responding speed to a test entry.

Table IV shows the results of training speed for the single-structured RBFNs using ISO, ErrCor, PSO, and APSO (for the results shown in Table II), compared to the multi-structured RBFNs using MC-PSO and MC-APSO (for the results shown in Table III), respectively.

Table IV results tip the scale of the MC-PSO and MC-APSO methods over all other single-structured methods. The multi-column methods separate the dataset into subsets and train them in a parallel structure. The parallelism feature reduces the time of the training process, where the population algorithms already tend to be faster in most cases than gradient methods. Moreover, even when multi-column methods use the same number of particles and hidden units of PSO or APSO, the individual RBFNs tuning computations are way less than single-structured ones. That is due to the smaller number of instances in each subset compared to the non-chopped dataset in the single structure RBFN.

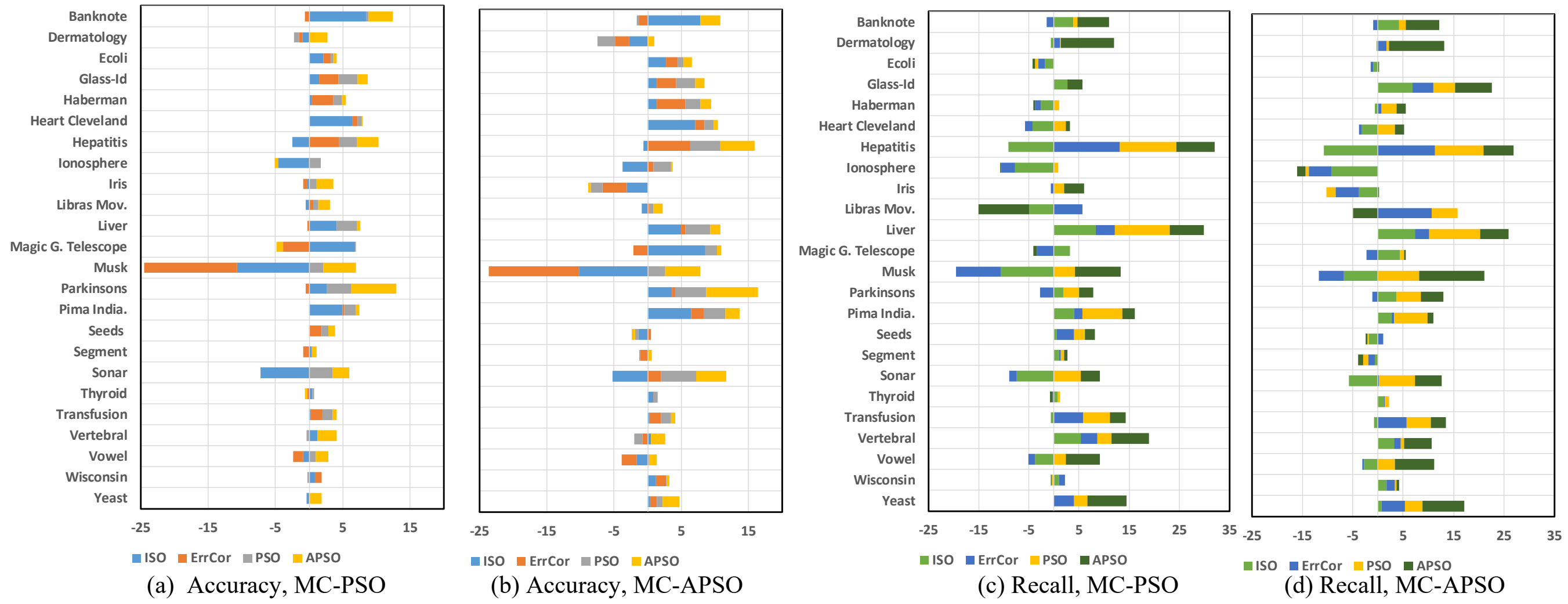

Fig. 5. The percentage differences in accuracy and recall for the non-preferable results of MC-PSO and MC-APCO in Table III, compared to the four traditional results of ISO, ErrCor, PSO and APSO shown in Table II.

TABLE IV

RESULTS OF TRAINING TIME (IN SECONDS) USING ISO, ERRCOR, PSO AND APSO ALGORITHMS FOR THE SINGLE-STRUCTURED RBFNS IN TABLE II, COMPARED TO THE MULTI-COLUMN RBFN USING MC-PSO AND MC-APSO ALGORITHMS IN TABLE III.

| Chopped Dataset | ISO | ErrCor | PSO | APSO | MC-PSO | MC-APSO |
|---|---|---|---|---|---|---|
| Banknote | 19.79±0.78 | 32.81±9.00 | 26.67±0.77 | 26.93±0.96 | **15.34±1.16** | 15.52±0.48 |
| Dermatology | 285.34±144.84 | 54.43±11.26 | 46.84±0.23 | 34.60±1.81 | **27.79±0.28** | 32.90±5.45 |
| Ecoli | 7.75±0.23 | 13.82±3.86 | 5.70±0.05 | 4.97±0.74 | **1.73±0.16** | 2.17±0.02 |
| Glass-Id | 5.94±0.28 | 6.91±2.51 | 2.57±0.02 | 2.56±0.02 | 1.90±0.14 | **1.86±0.29** |
| Haberman | 2.97±0.03 | 9.32±4.07 | 5.51±0.06 | 5.08±0.79 | **2.81±0.05** | 3.59±0.08 |
| Heart Cleveland | 11.36±0.25 | 27.34±0.43 | 7.31±0.05 | 7.02±1.34 | **3.77±0.79** | 4.03±0.04 |
| Hepatitis | 9.52±0.08 | 2.23±0.92 | 3.36±0.03 | 3.50±0.31 | **1.51±0.03** | 3.19±0.04 |
| Ionosphere | 137.72±0.49 | 24.01±9.36 | 10.21±0.05 | 9.98±0.02 | **4.02±0.06** | 5.15±0.07 |
| Iris | 2.08±0.72 | 3.83±0.88 | 5.75±0.05 | 5.64±0.13 | **1.88±0.13** | 2.08±0.11 |
| Libras Mov. | 369.71±467.59 | 11.65±0.11 | 8.29±0.34 | 9.87±0.12 | 4.65±0.02 | **4.53±0.29** |
| Liver | 6.06±0.04 | 15.49±0.90 | 8.86±0.10 | 7.61±1.30 | **1.75±0.13** | 2.52±0.49 |
| Magic G. Telescope | 549.66±1.12 | 1036.64±303.05 | 335.07±7.01 | 333.06±6.52 | 183.73±2.09 | **181.86±5.16** |
| Musk | 4600.55±38.57 | 642.36±246.08 | 11.87±0.08 | 11.29±0.29 | 9.91±1.17 | **9.70±1.11** |
| Parkinsons | 37.37±0.25 | 16.13±8.59 | 4.40±0.06 | 4.42±0.03 | **2.33±0.26** | 2.42±0.09 |
| Pima India. | 19.77±0.49 | 41.54±0.35 | 12.17±1.07 | 11.03±0.51 | **10.26±1.03** | 10.40±0.94 |
| Seeds | 14.24±0.47 | 14.66±3.35 | 5.49±0.79 | 5.20±0.27 | **4.55±0.18** | 7.82±0.67 |
| Segment | 122.42±0.68 | 26.45±8.19 | 40.81±0.10 | 41.13±0.15 | 22.84±0.11 | **22.48±0.30** |
| Sonar | 269.94±2.83 | 24.12±5.53 | 5.30±0.12 | 5.08±0.32 | 2.74±0.05 | **2.72±0.01** |
| Thyroid | 1155.54±7.14 | 636.82±5.61 | 364.12±1.64 | 364.74±4.18 | **219.09±21.22** | 254.40±1.13 |
| Transfusion | 11.25±0.12 | 32.82±0.23 | 12.71±0.04 | 11.68±1.04 | 6.08±0.14 | **4.54±0.76** |
| Vertebral | 18.98±0.60 | 22.90±15.23 | 10.38±0.08 | 10.26±0.04 | **9.45±2.03** | 10.12±1.71 |
| Vowel | 16.18±0.69 | 45.63±0.71 | 20.05±3.40 | 18.02±1.73 | **8.37±0.09** | 11.14±1.13 |
| Wisconsin | 29.14±0.38 | 55.17±6.64 | 14.62±1.26 | 14.89±1.19 | 3.81±0.05 | **3.67±0.38** |
| Yeast | 36.71±0.63 | 59.93±25.25 | 18.65±0.95 | 18.21±0.64 | **3.80±0.02** | 11.96±0.33 |

The ISO is always faster than the ErrCor due to the incremental nature of the latter. In most cases, the single-structured population methods: PSO and APSO, are more rapid than ISO and ErrCor. In few experimental cases, ISO reaches convergence faster than other single-structured population methods, while ISO is slow in others.

Testing speed is another important key factor in a successful NN model. It depends mainly on the complexity and memory usage of the trained NN structure and the internal runtime computations. The RBFN testing speed depends on the number of hidden units, the hidden unit vector dimensionality, and the complexity of the activation function. As previously mentioned, we intentionally kept the number of hidden units fixed for all our single-structured experiments. We set a maximum number of hidden units for our proposed method to compare similar ones and to see the effect of reducing others while improving accuracy and recall results.

Table V shows the testing speed results using the single-structured pre-trained RBFNs for ISO, ErrCor, PSO, or APSO (structured in Table II). Results are compared to the multi-column RBFNs with pre-trained MC-PSO and MC-APSO algorithms (structured in Table III). There is no difference in test speed for all four single-structured RBFNs. Therefore, only one-column results are shown in Table IV for single-structured RBFN. While the results differ from one dataset experiment to another since the number of features is different.

The MC-PSO and MC-APSO, in most cases, are faster than the other four single-structured RBFNs algorithms. With the same number of hidden units, the speed is almost similar, and the speed difference is not noticeable, because, at a critical parallel environment, the speed cannot be measured precisely due to the processing delay and mutual process waiting in the operating system tasks. However, the speed difference is huge for a smaller number of hidden units per individual RBFNs, and the speed significantly doubles with improved accuracy and recall results.

One can notice that using fewer hidden units in the multi-column structure per subset increases the total number of swarm particles and the total number of hidden units in the multi-column structure. For example, with five hidden units and 15 swarm particles per individual RBFN in the Ecoli dataset experiment, the total number of hidden units for the two subsets is $5 \times 2 = 10$, and the total number of particles $15 \times 2 = 30$. Compare that to the single structure experiment with only nine hidden units and 25 particles. The novel methods efficiently use the parallel environment and compact the number of instances to half per subset while using more total particles and hidden units. The MC-PSO method in the Ecoli experiment improves the accuracy results by 0.29%, the training speed by 3.29 times, and the testing speed by 2.75 times over PSO.

### *E. General Comparisons of the MC-PSO and MC-APSO Methods to evolutionary and MCRN Methods*

This subsection demonstrates accuracy and training time comparison for the two novel methods with some evolutionary techniques and the parallel MCRN method, as summarized in Table VI. The accuracy and training time results of ISO, ErrCor, PSO, APSO, MCRN, MC-PSO, and MC-APSO are collected from our experiments, as shown in Table VI. On the other hand, the results of Differential Evolution (DE) [40], Biogeography-Based Optimizer (BBO) [40], Ant Colony Optimization (ACO)[40], FireFly algorithm (FireFly)[40], and Multidimensional PSO (MD-PSO) [41] are the best results provided by the researchers in their corresponding papers.

Table VI accuracy results show a noticeable lead for the multi-column methods in general, where the multi-column uses specialized individual RBFNs trained with a subset of the dataset. MC-PSO and MC-APSO show win cases of 5/10 in accuracy results and show promising results compared to the winner of the rest results.

TABLE V
RESULTS OF TESTING TIME USING THE SINGLE-STRUCTURED RBFNS USING EITHER ISO, ERRCOR, PSO OR APSO ALGORITHMS FOR RBFN STRUCTURED SHOWN IN TABLE II, COMPARED TO THE MULTI-COLUMN RBFNS USING MC-PSO AND MC-APSO ALGORITHMS SHOWN IN TABLE III. (TIME IS IN SECONDS).

| Chopped Dataset | *H* | Single-structured RBF | *H* | MC-PSO | *H* | MC-APSO |
|---|---|---|---|---|---|---|
| Banknote | 9 | 0.0034±0.0017 | 9 | **0.0032±0.0009** | 9 | 0.0034±0.0009 |
| Dermatology | 9 | 0.0051±0.0037 | 9 | 0.0037±0.0002 | 9 | **0.0036±0.0001** |
| Ecoli | 9 | 0.0022±0.0019 | 5 | **0.0008±0.0001** | 5 | **0.0008±0.0001** |
| Glass-Id | 9 | 0.0019±0.0017 | 7 | **0.0006±0.0001** | 7 | **0.0006±0.0001** |
| Haberman | 9 | 0.0020±0.0017 | 7 | 0.0012±0.0009 | 9 | **0.0012±0.0008** |
| Heart Cleveland | 9 | 0.0020±0.0017 | 7 | **0.0008±0.0001** | 7 | **0.0008±0.0001** |
| Hepatitis | 9 | 0.0019±0.0017 | 5 | **0.0006±0.0000** | 9 | 0.0010±0.0006 |
| Ionosphere | 9 | 0.0022±0.0017 | 5 | **0.0009±0.0001** | 7 | **0.0009±0.0001** |
| Iris | 9 | 0.0027±0.0032 | 5 | 0.0009±0.0001 | 5 | **0.0009±0.0000** |
| Libras Mov. | 9 | 0.0024±0.0020 | 7 | **0.0012±0.0001** | 7 | **0.0012±0.0001** |
| Liver | 9 | 0.0020±0.0016 | 5 | 0.0008±0.0001 | 5 | **0.0007±0.0000** |
| Magic G. Telescope | 9 | **0.0266±0.0026** | 9 | 0.1019±0.0084 | 9 | 0.1024±0.0018 |
| Musk | 9 | 0.0028±0.0019 | 9 | 0.0025±0.0000 | 9 | **0.0024±0.0001** |
| Parkinsons | 9 | 0.0020±0.0018 | 9 | **0.0006±0.0000** | 9 | 0.0007±0.0000 |
| Pima India. | 9 | 0.0025±0.0016 | 9 | **0.0018±0.0000** | 9 | 0.0018±0.0001 |
| Seeds | 9 | 0.0026±0.0025 | 5 | **0.0011±0.0001** | 7 | 0.0013±0.0001 |
| Segment | 9 | **0.0049±0.0019** | 9 | 0.0074±0.0002 | 9 | 0.0087±0.0032 |
| Sonar | 9 | 0.0022±0.0021 | 9 | 0.0008±0.0001 | 9 | **0.0007±0.0001** |
| Thyroid | 9 | **0.0308±0.0026** | 9 | 0.0769±0.0016 | 9 | 0.0785±0.0031 |
| Transfusion | 9 | 0.0026±0.0017 | 9 | 0.0020±0.0009 | 7 | **0.0018±0.0009** |
| Vertebral | 9 | 0.0025±0.0017 | 7 | **0.0017±0.0001** | 7 | **0.0017±0.0001** |
| Vowel | 9 | 0.0030±0.0025 | 9 | **0.0024±0.0001** | 9 | **0.0024±0.0001** |
| Wisconsin | 9 | 0.0029±0.0017 | 5 | **0.0013±0.0001** | 5 | **0.0013±0.0001** |
| Yeast | 9 | 0.0037±0.0020 | 5 | **0.0032±0.0001** | 9 | 0.0042±0.0006 |

For the training time comparison, the MC-PSO and MC-APSO win all the experimental cases due to the parallelism technique used in their structure. Although MCRN uses the same parallelism technique, MCRN uses the slow ErrCor method as its internal RBFN training algorithm. The ErrCor method reduces the number of used hidden units and improves accuracy results. However, it shows slower training time compared to other methods (Table VI).

## V. CONCLUSION

The MC-PSO and MC-APSO show better accuracy results and faster training and testing speed compared to gradient methods: ISO, ErrCor; and compared to population methods PSO and APSO. The maximum improvement in MC-PSO accuracy reaches $8.45\%$, $4.43\%$, $3.63\%$, and $6.74\%$ over ISO in Banknote, ErrCor in Hepatitis, PSO in Parkinsons, and APSO in Parkinsons, respectively. Similarly, MC-APSO showed maximum accuracy improvements of $8.53\%$, $6.31\%$, $5.31\%$, and $7.71\%$ over ISO in Magic G. Telescope, ErrCor in Hepatitis, PSO in Sonar, and APSO in Parkinsons, respectively. On the other hand, the MC-PSO shows a maximum training speed improvement of $5.06$ times over its counterpart, the PSO method in Liver dataset. Likewise, MC-APSO shows a maximum training speed improvement of $4.06$ times over the APSO in Wisconsin dataset. In a one-vs-one comparison, MC-PSO is faster than MC-APSO due to the adaptation feature in the latter, while MC-APSO shows better accuracy results.

Compared to other evolutionary RBFN tuning methods such as DE, BBO, and MD-PSO, the MC-PSO shows a maximum accuracy improvement of $13.42\%$, $2.77\%$, and $25.38\%$, respectively. The MC-APSO shows a maximum improvement of $14.29\%$, $2.77\%$, and $24.67\%$, respectively. MC-PSO and MC-APSO show maximum accuracy over MCRN of $2.4\%$ and $2.36\%$.

The MC-PSO and MC-APSO show impressive training speed improvement over their counterpart, the MCRN. The MC-PSO shows a maximum improvement of $24.18$ times, while the MC-APSO shows a maximum improvement of $18.19$ times over the MCRN.

Our future plan is to consider applying neural network to recent well-performing PSO variants with large-scale [42], dynamic [43], and many-objective optimization [44]. As optimization of neural network has such complex characteristics and these PSO variants should be promising. Another future plan is to integrate our novel technique in recent deep learning technique as alternative to the fully connected layer of Convolutional networks.

TABLE VI

GENERAL COMPARISONS OF THE ACCURACY AND THE TRAINING TIME RESULTS OF THE NEW MC-PSO AND MC-APSO METHODS WITH TRADITIONAL, EVOLUTIONARY AND MULTICOLUMN TUNING METHODS

| Algorithm | Ecoli | Glass-Id | Heart Cleveland | Liver | Parkinsons | Pima India. | Thyroid | Transfusion | Vowel | Wisconsin |
|---|---|---|---|---|---|---|---|---|---|---|
| **Accuracy (Mean ±STD)** | | | | | | | | | | |
| ISO[5] | 94.64±1.69 | 92.51±3.50 | 76.76±4.09 | 65.40±6.42 | 84.61±7.91 | 70.31±0.95 | 95.48±0.36 | 79.95±2.49 | 98.48±0.51 | 96.15±2.33 |
| ErrCor[7] | 95.53±1.84 | 91.06±4.01 | 82.47±2.94 | 69.80±7.32 | 87.68±8.00 | 75.00±2.88 | 96.22±0.34 | 78.48±2.85 | **99.09±0.90** | 96.00±2.10 |
| PSO[10] | 96.43±2.49 | 91.07±3.90 | 82.47±3.60 | 66.59±6.88 | 83.60±3.83 | 73.56±5.11 | 95.69±0.08 | 78.75±1.98 | 96.57±1.31 | 97.29±1.61 |
| APSO[10] | 96.13±2.26 | 92.49±3.57 | 83.15±2.48 | 68.93±6.42 | 80.49±3.05 | 74.74±3.75 | 96.29±0.28 | 79.55±3.19 | 95.76±1.36 | 97.00±1.26 |
| DE[36]* | - | - | - | 56.10±2.32 | 74.03±6.72 | - | - | 76.63±0.33 | - | 86.26±9.17 |
| BBO[36]* | - | - | - | 69.15±4.36 | 86.42±1.92 | - | - | 77.45±0.28 | - | **97.86 ±0.37** |
| ACO[36]* | | | | 52.22±6.34 | 65.22±21.2 | | | 76.39±0.45 | | 68.53±21.0 |
| FireFly[36]* | | | | 61.17±2.36 | 79.55±3.38 | | | 77.22±0.43 | | 96.66±0.90 |
| MD-PSO[1]* | 87.20±0.50 | - | 59.90±1.30 | - | 85.70±1.20 | 75.60±0.90 | - | 78.20±0.70 | 72.20±1.70 | 95.90±0.70 |
| MCRN[4] | 96.11±2.04 | 91.55±5.20 | 82.81±4.43 | **72.44±5.61** | **90.25±5.56** | 76.69±3.43 | **96.79±0.30** | 79.28±2.48 | 95.67±0.52 | 97.29±1.36 |
| MC-PSO | 96.72±3.24 | **93.95±4.80** | 83.19±2.99 | 69.52±6.73 | 87.23±5.93 | 75.26±4.92 | 95.96±0.26 | **80.22±3.12** | 97.58±0.83 | 97.00±1.53 |
| MC-APSO | **97.32±2.44** | 93.91±2.67 | **83.82±3.79** | 70.39±6.05 | 88.20±6.69 | **76.82±2.17** | 96.29±0.35 | 80.22±2.40 | 96.87±0.66 | 97.43±1.70 |
| **Training Time (Mean ± STD)** | | | | | | | | | | |
| ISO[5] | 7.75±0.23 | 5.94±0.28 | 11.36±0.25 | 6.06±0.04 | 37.37±0.25 | 19.77±0.49 | 1155.54±7.14 | 11.25±0.12 | 16.18±0.69 | 29.14±0.38 |
| ErrCor[7] | 13.82±3.86 | 6.91±2.51 | 27.34±0.43 | 15.49±0.90 | 16.13±8.59 | 41.54±0.35 | 636.82±5.61 | 32.82±0.23 | 45.63±0.71 | 55.17±6.64 |
| PSO[10] | 5.70±0.05 | 2.57±0.02 | 7.31±0.05 | 8.86±0.10 | 4.40±0.06 | 12.17±1.07 | 364.12±1.64 | 12.71±0.04 | 20.05±3.40 | 14.62±1.26 |
| APSO[10] | 4.97±0.74 | 2.56±0.02 | 7.02±1.34 | 7.61±1.30 | 4.42±0.03 | 11.03±0.51 | 364.74±4.18 | 11.68±1.04 | 18.02±1.73 | 14.89±1.19 |
| MCRN[4] | 2.97±0.67 | 3.17±1.08 | 9.18±0.34 | 4.38±3.89 | 12.32±7.16 | 24.44±0.71 | 1582.00±26.42 | 17.86±0.51 | 202.41±60.75 | 6.49±0.84 |
| MC-PSO | **1.73±0.16** | 1.90±0.14 | **3.77±0.79** | **1.75±0.13** | **2.33±0.26** | **10.26±1.03** | **219.09±21.22** | 6.08±0.14 | **8.37±0.09** | 3.81±0.05 |
| MC-APSO | 2.17±0.02 | **1.86±0.29** | 4.03±0.04 | 2.52±0.49 | 2.42±0.09 | 10.40±0.94 | 254.40±1.13 | **4.54±0.76** | 11.14±1.13 | **3.67±0.38** |

*Results of DE, BBO, ACO, FireFly and MD-PSO are reported from their corresponding papers.